\documentclass{article}
\usepackage[utf8]{inputenc}

\usepackage{amsfonts,amsmath,amssymb}
\usepackage{booktabs}
\usepackage[ruled,vlined]{algorithm2e}
\usepackage{amsmath}
\usepackage[hyperfootnotes=false]{hyperref}
\usepackage{capt-of}
\usepackage{subcaption}
\usepackage{comment}
\usepackage{booktabs}
\usepackage{verbatim}
\usepackage{pdflscape}
\usepackage{makecell}
\usepackage{graphicx}
\usepackage{subcaption}
\usepackage{multirow}
\usepackage{algpseudocode}
\usepackage{geometry}
\usepackage{adjustbox}
\usepackage{changepage} 
\usepackage[hypcap=false]{caption}
\usepackage{rotating}   
\usepackage{xcolor}

\usepackage{relsize}
\usepackage[symbol]{footmisc}

\begin{document}

{
\Large
\noindent\textbf{Integrating Statistical Significance and Discriminative Power\vskip 0.2cm\noindent in Pattern Discovery}
}
\vskip 0.4cm
\noindent Leonardo Alexandre\textsuperscript{1,2,*},
Rafael S. Costa\textsuperscript{2},
Rui Henriques\textsuperscript{1,*}
\vskip 0.4cm
\noindent \textbf{\textsuperscript{1}} INESC-ID and Instituto Superior T\'{e}cnico, Universidade de Lisboa, Lisboa, Portugal
\\
\textbf{\textsuperscript{2}} LAQV-REQUIMTE, Department of Chemistry, NOVA School of Science and Technology, Universidade NOVA de Lisboa, 2829-516 Caparica, Portugal
\textcolor{white}{\footnote[1]{Correspondence \textsf{\{leonardoalexandre,rmch\}@tecnico.ulisboa.pt}}}

\section*{Abstract}
\vskip -0.2cm
Pattern discovery plays a central role in both descriptive and predictive tasks across multiple domains. Actionable patterns must meet rigorous statistical significance criteria and, in the presence of target variables, further uphold discriminative power. Our work addresses the underexplored area of guiding pattern discovery by integrating statistical significance and discriminative power criteria into state-of-the-art algorithms while preserving pattern quality. We also address how pattern quality thresholds, imposed by some algorithms, can be rectified to accommodate these additional criteria. To test the proposed methodology, we select the triclustering task as the guiding pattern discovery case and extend well-known greedy and multi-objective optimization triclustering algorithms, $\delta$-Trimax and TriGen, that use various pattern quality criteria, such as Mean Squared Residual (MSR), Least Squared Lines (LSL), and Multi Slope Measure (MSL). Results from three case studies show the role of the proposed methodology in discovering patterns with pronounced improvements of discriminative power and statistical significance without quality deterioration, highlighting its importance in supervisedly guiding the search.  Although the proposed methodology is motivated over multivariate time series data, it can be straightforwardly extended to pattern discovery tasks involving multivariate, N-way (N$>$3), transactional, and sequential data structures.\vskip 0.2cm

\noindent\textbf{Availability:} The code is freely available at \href{https://github.com/JupitersMight/MOF_Triclustering}{https://github.com/JupitersMight/MOF\_Triclustering} under the MIT license.

\section{Introduction}
\vskip -0.2cm
Pattern mining is a widely researched field of data analysis that has served as the backbone of decision-making across many fields of expertise \cite{xie2019time}. In this context, knowledge acquisition is driven by the discovery of unexpectedly frequent substructures from a given data structure. In tabular data, pattern discovery generally aims to find coherent subspaces defined by subsets of observations and variables \cite{madeira2004biclustering}. When handling more complex data structures (e.g. cubic data, three-way data) triclustering algorithms define coherent subspaces across the three dimensions, typically referred to as sub-cubes or triclusters \cite{henriques2018triclustering}. Finding coherent subspaces depends on the merit function implemented by the algorithm to guide the search. Typically, algorithms ensure that extracted patterns (1) are dissimilar, (2) yield statistical significance \textit{e.g.} an unexpectedly low probability of occurrence against a null data model, and (3) meet rigorous quality criteria \cite{henriques2018triclustering}. Additionally, in the presence of annotated data, patterns can be assessed using interestingness measurements \cite{alexandre2022disa}. Patterns exhibiting a high discriminative power towards an outcome of interest play a central role in enhancing predictive tasks \cite{alexandre2021mining, soares2022learning}.

Current research in the field of discriminative pattern mining aims at improving the efficiency of the searches \cite{seyfi2023h}, reducing false positive discoveries \cite{henriques2018bsig}, promoting actionability \cite{pellegrina2019spumante}, and using discriminative patterns to guide predictive approaches \cite{henriques2021flebic}. Complementarily, statistical tests for assessing the statistical significance of patterns have been considered for transactional data structures \cite{tang2022mining}. In this context, statistical significance criteria are applied to the pattern discriminative power and not the pattern itself \cite{tang2022mining}. Additionally, others in the field focus on capturing different types of correlation \cite{gutierrez2014lsl,gutierrez2015msl}, improving algorithms to handle different data types \cite{chen2016computational} and their scalability \cite{melgar2021discovering}. To the best of our knowledge, there are no approaches that integrate quality criteria, discriminative criteria, and statistical significance to guide pattern discovery. In this context, we provide an in-depth discussion on how to properly incorporate \textit{statistical significance} and \textit{discriminative criteria} into the merit functions of state-of-the-art algorithms without forgoing pattern quality. Furthermore, we discuss how to automatically revise the quality thresholds associated with the pursued merit functions as imposed by some algorithms (e.g. $\delta$-Trimax \cite{bhar2013coexpression}).

Considering the increasing ability to monitor the behavior of systems along time, a notable rise in the availability of tensor data structures, particularly three-way data, has been observed. The intrinsic properties of three-way data play a significant role in the understanding of underlying processes and relationships, rendering it a valuable resource for pattern-guided knowledge acquisition and decision support. To this end, we modify two well-known greedy-based triclustering algorithms for temporal pattern discovery in three-way data, $\delta$-Trimax\cite{bhar2013coexpression} and TriGen\cite{gutierrez2014trigen}, and assess the proposed methodology in the context of three case studies. We observe that the algorithms produce significant improvements in both the \textit{discriminative power} and \textit{statistical significance} of the formed patterns. The proposed methodology is not algorithm-constrained nor data-constrained, thus it can be extended and used with any $N$-dimensional pattern mining algorithm.  

The paper is organized as follows: the remainder of this section provides the essential background. 
Section 2 introduces the proposed methodology. Section 3 discusses the gathered results from real-world and synthetic data. Concluding remarks and implications are finally drawn.

\subsection{Background}

Consider a two-dimensional dataset (matrix data), denoted as \textbf{A}, defined by $n$ observations ($X = {x_1, ..., x_n}$) and $m$ variables ($Y = {y_1, ..., y_m}$), featuring $n \times m$ elements $a_{ij}$. The tasks of \textit{clustering} and \textit{biclustering} aim to extract \textit{clusters} $I_i \subseteq X$ or \textit{biclusters} $B_k = (I_k, J_k)$, each represented by a subset of observations $I_k \subseteq X$ correlated on a subset of variables $J_k \subseteq Y$. \textbf{Homogeneity} criteria, typically ensured through the application of a merit function, guide the formation of (bi)clusters \cite{madeira2004biclustering}. For a given bicluster, its values $a_{ij}=c_j+\gamma_i+\eta_{ij}$ can be explained by value expectations $c_j$, adjustments $\gamma_i$, and noise $\eta_{ij}$. If each element of \textbf{A} corresponds to a bicluster, the \textbf{bicluster pattern} $\varphi_{B}$ is the ordered set of values in the absence of adjustments and noise, expressed as $\varphi_{B} = \{ c_{j} | y_j \in J\}$. In addition to the homogeneity criteria, criteria for \textbf{\textit{statistical significance}} \cite{henriques2018bsig} ensure that the probability of a bicluster's occurrence (compared to a null model) deviates from expectations. Moreover, \textbf{dissimilarity} criteria \cite{henriques2017bicpams} can be applied to further ensure the absence of redundant biclusters.

A three-dimensional dataset (three-way tensor data), denoted as \textbf{A}, is characterized by $n$ observations $X = \{x_1, ..., x_n\}$, $m$ variables $Y = \{y_1, ..., y_m\}$, and $p$ contexts $Z = \{z_1, ..., z_p\}$. The elements $a_{ijk}$ establish a connection between observation $x_i$, attribute $y_j$, and context $z_k$. Similar to two-dimensional data, three-dimensional data can be \textit{real-valued} $(a_{ijk} \in \mathbb{R})$, symbolic ($a_{ijk} \in \Sigma$, where $\Sigma$ is a set of nominal or ordinal symbols), integer ($a_{ijk} \in \mathbb{Z}$), or \textit{non-identically distributed} ($a_{ijk} \in \mathcal{A}_j$, where $\mathcal{A}j$ is the domain of variable $y_j$), and may contain missing elements, $a_{ijk} \in \mathcal{A}_j \cup \emptyset$. When the context dimension represents a set of time points ($Z = \{t_1, t_2, ..., t_p\}$), we encounter a temporal three-dimensional dataset, also known as $m$-order multivariate time series data.

Given three-way tensor data \textbf{A} with $n$ observations, $m$ variables, and $p$ contexts/time points, a \textbf{tricluster} $T = (I, J, K)$ represents a subspace of the original space, where $I \subseteq X$, $J \subseteq Y$, and $K \subseteq Z$ are subsets of observations, variables, and contexts/time points, respectively \cite{henriques2018triclustering}. The \textbf{triclustering task} aims to discover a set of triclusters $\mathcal{T} = \{T_1, ..., T_l\}$ such that each tricluster $T_i$ satisfies specific criteria of homogeneity, dissimilarity, and \textit{statistical significance}. Each element of a tricluster, $a_{ijk}$, is described by a base value $c_{jk}$, observation adjustment $\gamma_i$, and noise $\eta_{ijk}$. The tricluster \textbf{pattern}, $\varphi_{T}$, constitutes an ordered set of value expectations along the subset of variables and context dimensions in the absence of adjustments and noise: $\varphi_{T} = \{c_{jk} | y_j \in J, z_k \in K, c_{jk} \in Y_k\}$. The coverage $\Phi$ of the tricluster pattern $\varphi_{T}$, denoted as $\Phi(\varphi_{T})$, corresponds to the number of observations containing the tricluster pattern $\varphi_{T}$. This same logic can be extended to a nominal outcome of interest $c$, where $c$ can take any value in the class variable. The coverage of the outcome, defined as $\Phi(c)$, represents the number of observations with the outcome of interest. For instance, considering the class variable $y_{out} = [a, b, a, a]$, the coverage $\Phi(a)$ is equal to $3/4$.

\textbf{Homogeneity} criteria play a crucial role in determining the structure, coherence, and quality of a triclustering solution \cite{henriques2018triclustering}. In this context:

\begin{itemize}
 \item[--]\vskip -0.3cm the \textit{structure} is defined by the number, size, shape, and position of triclusters;
 \item[--]\vskip -0.2cm the \textit{coherence} of a tricluster is defined by the observed correlation of values (coherence assumption) and the allowed deviation from expectations (coherence strength);
 \item[--]\vskip -0.2cm the \textit{quality} of a tricluster is defined by the type and amount of tolerated noise.
\end{itemize}
\vskip -0.2cm

A tricluster exhibits \textbf{constant coherence} when the subspace demonstrates constant values (for symbolic data) or approximately constant values (for real-valued data). 

\textbf{Dissimilarity} criteria ensure that any tricluster similar to another tricluster with higher priority is removed from $\mathcal{T}$ and possibly used to refine similar triclusters in $\mathcal{T}$. A tricluster $T = (I, J, K)$ is considered \textit{maximal} if and only if there is no other tricluster $(\textbf{I',J',K'})$ such that $I \subseteq I' \wedge J \subseteq J' \wedge K \subseteq K'$, satisfying the given criteria. 

Foremost, \textbf{\textit{statistical significance} criteria} ensure that the probability of each retrieved tricluster occurring against a null data model is unexpectedly low. Given the probability of the occurrence of a tricluster pattern, $p_{\varphi_{T}}$, binomial tails can be employed to robustly compute the probability of a tricluster $T = (I, J, K)$ with pattern $\varphi_{T}$ occurring for $|I|$ or more observations \cite{henriques2018bsig}. Accordingly,


\begin{equation}
\text{\textit{p}-value} =  P(x\ge |I| \mid x\sim Bin(p_{\varphi_{T}}, |X|)) = \sum_{x=|I|}^{|X|} {|X| \choose x} p_{\varphi_{T}}^x (1-p_{\varphi_{T}})^{|X| - x},
\end{equation}

\noindent where $Bin(p_{\varphi_{T}},|X|)$ is the null binomial distribution, $p_{\varphi_{T}}$ the tricluster pattern probability, $|X|$ the sample size, and $|I|$ the number of successes. A more detailed formulation can be found at \cite{alexandre2023trisig}.

The \textbf{discriminative properties} of a tricluster pattern $\varphi_{T}$, along with other interestingness measurements, can be evaluated using association rules \cite{alexandre2022disa}. Association rules depict a connection between two events, defined by a left-hand side (antecedent) and a right-hand side (consequent). In this context, an association rule may take the form $\varphi_{T} \to c$, where a pattern in the antecedent discriminates an outcome of interest in the consequent. The coverage of the association rule, $\Phi(\varphi_{T} \to c)$, is determined by the number of observations where both the pattern $\varphi_{T}$ and the outcome $c$ co-occur.

Two established interestingness measures are the \textit{confidence}, $\Phi(\varphi_{T} \to c) / \Phi(\varphi_{T})$, assessing the probability of $c$ occurring when $\varphi_{T}$ occurs, and the \textit{lift}, $(\Phi(\varphi_{T} \to c) / (\Phi(\varphi_{T}) \times \Phi(c)) \times N$, which further considers the probability of the consequent to evaluate the dependence between the consequent and antecedent. Additionally, a standardized version of the lift measure \cite{mcnicholas2008standardising} has proven to be particularly useful as a complementary measurement. In contrast to the lift measure, the standardized lift measure returns values between $[0;1]$, indicating whether the pattern has achieved its maximum lift, with a value of $1$ implying it has.


\begin{equation}
 standard\ lift = \frac{lift - \omega}{v - \omega},
\end{equation}

\noindent where $\omega$ is defined as:

\begin{equation}
 \omega = \max \left (\ \frac{1}{\Phi(\varphi_{T})} + \frac{1}{\Phi(c)} -1,\ \frac{(1 / |X|)}{ (1/\Phi(\varphi_{T})) \times (1/ \Phi(c))}\ \right),
\end{equation}

\noindent and $v$ is defined as:

\begin{equation}
 v = \frac{1}{\max (1/\Phi(\varphi_{T}),  1/\Phi(c))}.
\end{equation}

Triclustering algorithmic approaches belong to two major categories: 1) greedy or stochastic, or, 2) (quasi-)exhaustive   \cite{henriques2018triclustering}. Each approach can be further categorized according to the underlying behavior that it relies on, such as biclustering-based searches, and evolutionary multiobjective optimization algorithms, among others. Considering the scalability limits of exhaustive searches, this work places a focus on greedy-based approaches. Greedy algorithms follow two major approaches: 1) initialize a big pattern and iteratively refine it based on quality criteria until a satisfactory solution is obtained\cite{bhar2013coexpression}, or 2) initialize multiple small patterns and iteratively merge them based on quality criteria until a maximum predefined number of iterations is reached\cite{gutierrez2014trigen}. The latter approach is usually employed by evolutionary algorithms \cite{mukhopadhyay2015survey}. Irrespectively of the selected approach, the extracted patterns will reflect a coherent temporal progression with a specific structure and quality, that represent the \textit{homogeneity} of the pattern. Largely applied triclustering algorithms, including \textit{MOGA3C} \cite{liu2008multi} and $\delta$-Trimax \cite{bhar2013coexpression}, focus on optimizing each pattern with regards to pattern size, or pattern quality (e.g. mean squared residual score\cite{cheng2000biclustering}). Others in the field have focused their efforts on developing new merit functions or extending already well-established ones to be used in more complex data structures. Kakati \textit{et al.} \cite{kakati2018thd}, implements the Shifting-and-Scaling Similarity measure (SSSim) \cite{ahmed2014shifting} in the context of tensor data. Similarly, both Bhar \textit{et al.} \cite{bhar2013coexpression} and Gutiérrez-Avilés and Rubio-Escudero \cite{gutierrez2014mining}, proposed an extension of the mean square residue score (MSR) \cite{cheng2000biclustering} to be incorporated into merit functions of triclustering algorithms. Gutiérrez-Avilés and Rubio-Escudero \cite{gutierrez2014lsl, gutierrez2015msl} proposed two evaluation measures for triclusters, termed Least Squared Lines (LSL) and Multi Slope Measure (MSL), that focus on the similarity among the angles of the slopes formed by each profile of a tricluster pattern. 
Both LSL and MSL are embedded in merit functions employed by TriGen \cite{gutierrez2014trigen}, a multi-objective evolutionary algorithm, and its' extension STriGen \cite{melgar2021discovering}. Merit functions have also been tailored to fit domain-specific problems. Amaro-Mellado \textit{et al.} \cite{amaro2021generating} extracted seismic source zones by modifying TriGen merit functions to accommodate the geophysical indicators. In the presence of gene expression data, Gutiérrez-Avilés and Rubio-Escudero \cite{gutierrez2016triq}, proposed the addition of biological quality to merit functions (functional enrichment of the extracted regulatory patterns \cite{gene25ashburner}).

Two well-established greedy-based triclustering algorithms are: 1) $delta$-Trimax \cite{bhar2013coexpression}, a greedy iterative search algorithm that finds maximal triclusters having a score below a user-defined threshold ($\delta$), and, 2) TriGen \cite{gutierrez2014trigen}, an evolutionary multiobjective algorithm. Evolutionary algorithms have three main steps: 1) an initialization step that creates the initial population, 2) an evaluation step to assess the quality of each individual, given by a fitness function, to decide which individuals survive for the next iteration, and, 3) crossover and mutation, where individuals form connections and perform changes that will pass onto future generations (for more detailed formulations, the reader can check \cite{gutierrez2014trigen}). Despite the two aforementioned algorithms having different underlying behaviors to extract patterns, both implement well-established objective functions. The $\delta$-Trimax utilizes the mean square residue score (MSR) described by:

\begin{equation}
 MSR = \frac{1}{|I|\times|J|\times|K|} \mathlarger{\mathlarger{\sum}}_{i \in I, j \in J, k \in K}^{} (a_{ijk} - a_{iJK} - a_{IjK} - a_{IJk} + 2 \times a_{IJK})^2,
\end{equation}

\noindent where $a_{iJK}$, $a_{IjK}$, and $a_{IJk}$, correspond to the average of values for observation $\vec{x}_i$ given the subset of variables $J$ and contexts $K$, average of values for variable $\vec{y}_j$ given the subset of observations $I$ and contexts $K$, and average of values for context $\vec{z}_k$ given the subset of observations $I$ and variables $J$ respectively.

Trigen implements a wide variety of objective functions, of which we will focus on MSR, multi-slope measure (MSL), and least-squared line (LSL). Both LSL and MSL focus on graphical similarity. LSL measures the similitude between the least squares approximation for the points in each graphic of the three views, described by the following equation:

\begin{equation}
 LSL = \frac{AC_{lsl}(\{I,J,K\}) + AC_{lsl}(\{I,K,J\}) + AC_{lsl}(\{K,I,J\})}{3},
\end{equation}

\noindent where $AC_{lsl}(\{I,J,K\})$, $AC_{lsl}(\{I,K,J\})$, and $AC_{lsl}(\{K,I,J\})$ represent the least squares approximation operation over the graphical representation of the tricluster. 




MSL measures the similarity among the angles of the slopes formed by each profile, 

\begin{equation}
 MSL = \frac{AC_{multi}(\{I,J,K\}) + AC_{multi}(\{I,K,J\}) + AC_{multi}(\{K,I,J\})}{3},
\end{equation}

\noindent where $AC_{multi}(\{I,J,K\})$, $AC_{multi}(\{I,K,J\})$, and $AC_{multi}(\{K,I,J\})$ represent a multiangular comparison operation over the graphical representation of the tricluster. Due to the detailed extent of each component, we refer the reader to \cite{gutierrez2014trigen,gutierrez2015msl} for a more detailed formulation.

\section{Methodology}


The proposed methodology offers principles on how to extend objective functions to accommodate both \textit{statistical significance} views and \textit{discriminative criteria} without compromising pattern quality. To this end, we propose the creation of two components based on well-established interestingness measures in the presence of annotations \cite{alexandre2022disa} - \textit{discriminative power} component (DPC) - and on sound statistical tests \cite{alexandre2023trisig} - \textit{statistical significance} component (SSC).



\begin{figure}[!t]
\centering
\scriptsize
\includegraphics[width=\textwidth]{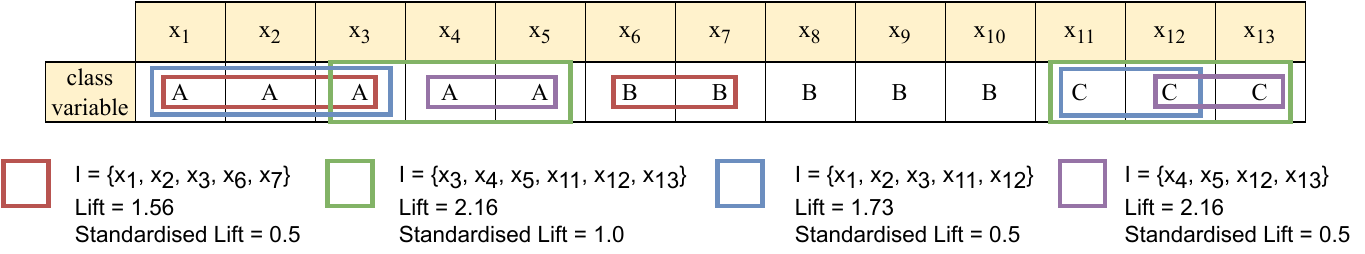}
\caption{\small Example of standard lift and lift measurements in the context of four patterns and one class variable with three outcomes. The red pattern discriminates outcome A with a lift of $1.56$ and standardized lift of $0.5$, the blue pattern discriminates outcome C with a lift of $1.73$ and standardized lift of $0.5$, the green pattern discriminates outcome C with a lift of $2.16$ and standardized lift of $1$, the purple pattern discriminates outcome C with a lift of $2.16$ and standardized lift of $0.5$.}
\label{standard_lift_example}
\end{figure}

By default, DPC incorporates two measures, the lift measure, and its' standardized version \cite{mcnicholas2008standardising},

\begin{equation}
 DPC = w_{d1} \times \ normalized \ lift + w_{d2} \times \ standard \ lift,
\end{equation}

\noindent where $w_{d1}$ and $w_{d2}$ represent user-defined weights, and $normalized\ lift$  - defined as \textit{desired lift}/\textit{lift}, if $lift > desired\ lift$, and 1 otherwise.  By default, both $w_{d1}$ and $w_{d2}$ have the same value but can be adjusted depending on user preferences for finding patterns with a higher \textit{discriminative power} ($w_{d2}$) versus maximizing the \textit{discriminative power} of each pattern ($w_{d1}$). Figure \ref{standard_lift_example} provides an example of how patterns with the same discriminative power vary in terms of completeness and vice versa. Additionally, the user can also adjust the minimum \textit{lift} necessary for a pattern to be discriminative - given by the \textit{desired lift} variable. By default, \textit{desired lift} has a value of $1.2$, but any value above $1.0$ already forces the formation of patterns with discriminative properties.

The SSC implements the \textit{statistical significance} tests proposed by Alexandre \textit{et al.} \cite{alexandre2023trisig}, but the use of other significance tests is also supported. A significance threshold ($\theta$) of $<0.05$, where we test the probability of each pattern against a null data model, is set as a standard threshold to inhibit false positives. Depending on the algorithmic approach (e.g. if we know the number of patterns before the search), further adjustments to this threshold can be done via the Bonferroni Correction \cite{henriques2018bsig}. 
To allow for a less abrupt transition when evaluating \textit{statistical significance}, we adjust the significance component by scaling the p-values with a logarithmic function. Given the aforesaid, the SSC is then defined as:


\begin{equation}
 SSC = \begin{cases}
 \frac{1}{|\log (\text{\textit{p}-value})|}, & \text{if \text{\textit{p}-value}} < \theta\\
 1, & \text{otherwise}.
  \end{cases}
\end{equation}

Finally, to embed the proposed components into existing objective functions we consider two main approaches, one based on addition and one on multiplication. The choice depends on how we choose to view the composition of components. If we assume that it is linear, addition is a natural way of capturing that behavior. Additionally, by using addition it is easier to model the degree to which both the \textit{discriminative power} and \textit{statistical significance} components compensate for pattern quality. Multiplication is suggested to penalize the score of patterns more abruptly. Contrary to the additive approach, the changes in the values of one of the components produce a proportional impact on the outcome statistic. With this in mind, we define the modified objective function (MOF),

\begin{equation} \label{eq:1}
  MOF =\begin{cases}
 \beta _1\ PQC^{\alpha _1} + \beta _2\ (DPC \times PQC)^{\alpha _2} + \beta _3\ (SSC \times PQC)^{\alpha _3} , & \text{if additive,}\\
 \ PQC^{\alpha _1} \times \ DPC^{\alpha _2} \times \ SSC^{\alpha _3}, & \text{if multiplicative},
  \end{cases}
\end{equation}

\noindent where $\beta _{(1-3)}$ and $\alpha _{(1-3)}$ represent the weights and adjustments of each component. The default value for the $\beta$ weights is 1 and for the $\alpha$ adjustments is $1/k$ where $k$ is the number of components. Further adjustments can be made to better understand the reasoning behind the proposed default values recall that state-of-the-art algorithms follow one of two approaches, either bottom-up or top-down. For algorithms that follow a top-down approach with a strict PQC threshold, we need to consider how the incorporation of DPC and SSC modules impacts the imposed threshold. Therefore, we explore the behavior of the MOF when a constant PQC value is applied.

\begin{figure}[!b]
\begin{adjustwidth}{-2.7cm}{-1.8cm}
\begin{minipage}{1.35\textwidth}
\vskip -0.3cm
\centering
\scriptsize
\includegraphics[height=3.7in]{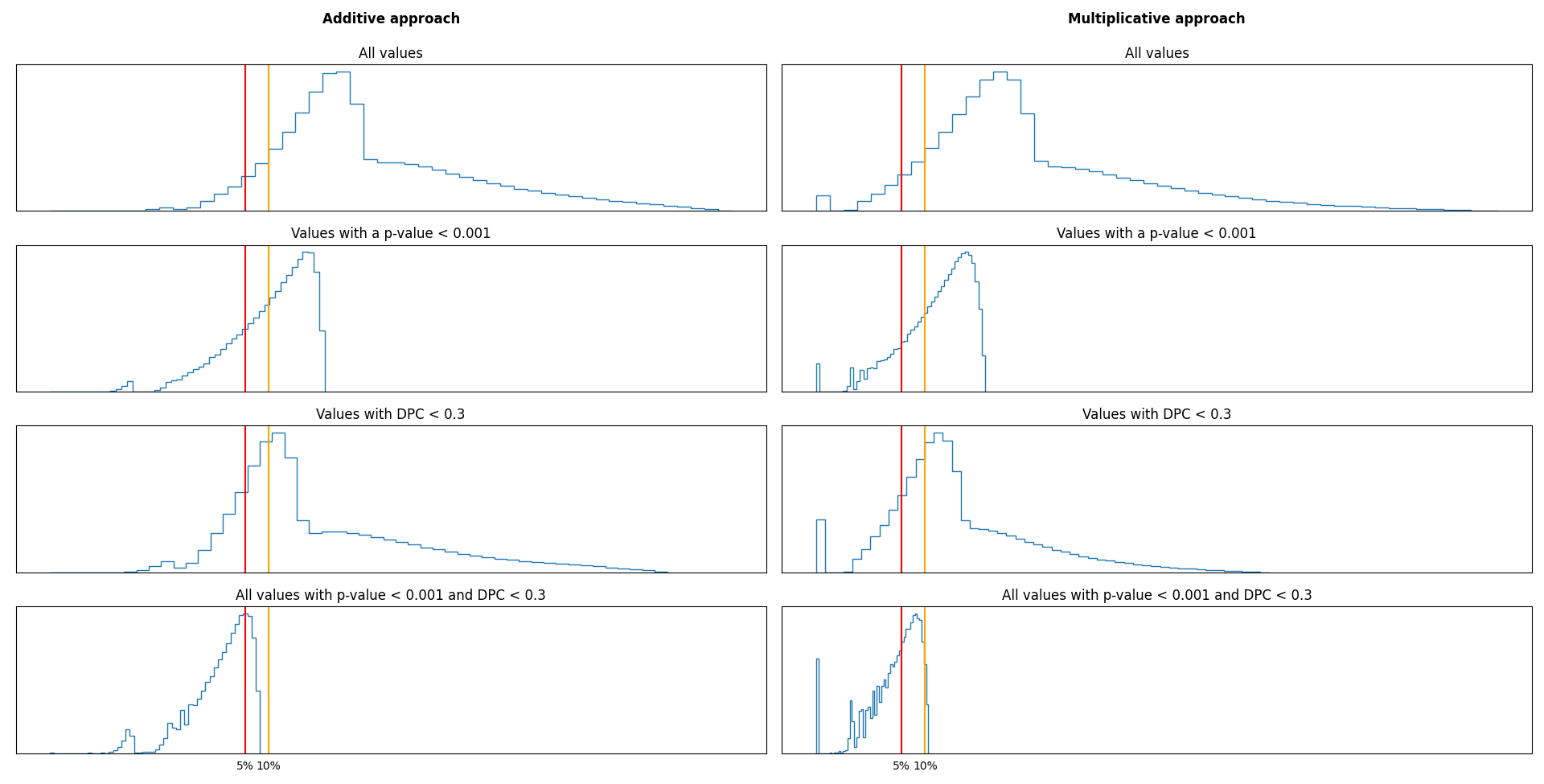}\\
\end{minipage}
\end{adjustwidth}

\caption{\small Illustration of the modified objective function based on additive approach, left column plots, multiplicative approach, right column plots. From top to bottom, each plot represents the distribution of values when considering: 1) all values, 2) values whose original \textit{p}-values are less than $0.001$, 3) values whose DPC has a value less than $0.3$, and, 4) values with the two aforesaid restrictions. The x-axis presents the value returned by the modified objective function and the y-axis is the frequency of each value. To generate the values for the distribution a fixed PQC is used, $DPC \sim U(0,\ 1)$ and \textit{p}-values are either sampled from $U(0,\ \theta)$ or $U(\theta,\ 1)$ with equal probability. Values to the left of the red line belong to the 5\% lowest values of the distribution. Similarly, the yellow line marks the value of the 10th percentile.}
\label{formula_distribution}
\end{figure}



Figure \ref{formula_distribution} illustrates how MOF varies for different DPC values and \textit{p}-values. We observe that in both approaches the distribution of values approximately follows a normal distribution, and as a result, we can examine binomial tails to study stricter and rarer MOF values. In this context, lower values translate into a better pattern solution, as such we concentrate on the right tail of the binomial distribution marked by the commonly applied boundaries of 5\%  and 10\%. By filtering out values using slightly stricter criteria for each component, we notice that almost all possible pattern solutions abiding by the criteria, fall under the 10\% right tail. Since the aforementioned MOF distribution is not molded by PQC values, we can use these boundaries to redefine the original user-inputted pattern quality threshold. The new threshold is then defined as $MOF_{\frac{m}{20}}$, where $MOF_{\frac{m}{20}}$ is the $\frac{m}{20}$th modified objective function value from an ordered array of values $\langle MOF_1, MOF_2, ..., MOF_m \rangle$, with size $m$. Each value is defined by equation \ref{eq:1} where $DPC \sim U(0,\ 1)$ and \textit{p}-values are either sampled from $U(0,\ \theta)$ or $U(\theta,\ 1)$ with equal probability.




\section{Computational Experimental Setup}

In order to assess the effectiveness of the proposed objective function in aiding the discovery of patterns with discriminative and statistical significance guarantees, we selected two greedy-based triclustering algorithms. Each algorithm represents one of the two major search approaches. First, $\delta$-Trimax \cite{bhar2015multiobjective}, an iterative search algorithm that follows a top-down approach. For a valid pattern solution, it imposes a strict MSR threshold that must be satisfied. Second, TriGen \cite{gutierrez2014trigen}, a multi-objective evolutionary algorithm that follows a bottom-up approach. It uses a fitness function based on pattern quality measures to guide pattern formation. 


The modified algorithms are then applied in the context of three publically available \textbf{datasets}: \textit{activity recognition} data \cite{misc_activity_recognition_system_based_on_multisensor_data_fusion}, \textit{basketball movements} data \cite{misc_basketball_dataset_587}, and \textit{sports activies} data \cite{misc_daily_and_sports_activities_256}. Table \ref{data_description} provides a brief description of each dataset and the applied preprocessing. In the case of the $\delta$-Trimax algorithm, the imposed MSR threshold is based on the dataset and also on the approach. For the \textit{activity recognition} data $\delta _{original} = 10^{-2}$, $\delta _{additive} = 10^{-3}$, and $\delta _{multiplicative} = 10^{-4}$, for the original algorithm, the algorithm modified with the additive approach, and the algorithm modified with a multiplicative approach, respectively. For the \textit{basketball movements} data $\delta _{original} = 10^{-3}$, $\delta _{additive} = 10^{-3}$, and $\delta _{multiplicative} = 10^{-4}$. Finally, for the \textit{sports activities} data, $\delta _{original} = 10^{-4}$, $\delta _{additive} = 10^{-4}$, and $\delta _{multiplicative} = 10^{-3}$. Regarding the TriGen parameter setup, we select a total of 20 clusters to be extracted with every other parameter set to their default values.


\begin{table}[!b]
\vskip -0.2cm
\caption{Summary of the selected case studies, including the size and domain of each dimension (e.g., sports, sensor data, time), and the applied preprocessing.}
\vskip -0.3cm
\begin{adjustwidth}{-1cm}{0cm}
\scriptsize
\begin{tabular}{p{3.6cm}p{9.8cm}p{2.5cm}}\toprule
Datasets & Description & Preprocessing \\ \midrule
\mbox{\textit{activity recognition}} \mbox{(activity x sensor data x time)} \mbox{(88 x 6 x 480)} & Real-life benchmark in the area of Activity Recognition to predict the activity performed by a user from time-series generated by a Wireless Sensor Network (WSN), according to the EvAAL competition technical annex (http://evaal.aaloa.org/). 
& \mbox{-- Feature scaling;} \\\midrule
\textit{basketball movements} \newline \mbox{(basketball movement} \mbox{x sensor data x time}) \mbox{(79 x 6 x 100)} & A total of 4 users were asked to do the following 5 activities: pass, hold the ball, shoot pick up the ball, and dribble. Data was gathered using accelerometer measures and gyroscope measures. & \mbox{-- Feature scaling;} \newline \mbox{-- Piecewise Aggregate} \newline \mbox{Approximation (PAA);} \newline \mbox{(100 time points)} \\\midrule
\mbox{\textit{sports activities}} \newline \mbox{(sport x sensor data x time)} \mbox{(152 x 46 x 375)} & The subjects were asked to perform the activities in their own style and were not restricted on how the activities should be performed. Sensor units were used to acquire data at 25 Hz sampling frequency. The 5-min signals are divided into 5-sec segments so that 480(=60x8) signal segments are obtained for each activity. 
& \mbox{-- Feature scaling;} \newline \mbox{-- Piecewise Aggregate} \newline \mbox{Approximation (PAA);} \newline \mbox{(375 time points)}\\\bottomrule
\end{tabular}
\label{data_description}
\end{adjustwidth}
\vskip -0.2cm
\end{table}

Finally, we assess pattern solutions returned by both $\delta$-Trimax and Trigen, and their modified versions, assessing them against each other. We use the t-student test to understand if there is a significant improvement on (1) pattern quality; (2) spearman and pearson correlation metrics, used to test improvements in similar instances \cite{gutierrez2014lsl, gutierrez2015msl}; and (3) discriminative and statistical significant criteria. Additionally, we provide the outcome distribution between the original algorithms and their modified versions. 

\section{Results and Discussion}

We hypothesize that the extension of triclustering algorithms, with the proposed modified objective functions, produce triclustering solutions where patterns exhibit an overall improvement in both \textit{discriminative power} and \textit{statistical significance} against traditional searches solely focused on homogeneity criteria. In Table \ref{tab_results} we present the results of this analysis for each case study and methodologic approach. Each pattern solution is assessed with regard to the average volume of patterns, pattern quality measurements, \textit{discriminative power} criteria, \textit{statistical significance criteria}, variable correlation criteria, and objective function values.



Considering the \textit{activity recognition} data, we observe an overall improvement of metric values across all pattern solutions using modified objective functions.  Specifically, the additive approach in the TriGen algorithm revealed significant improvements in the \textit{discriminative power} and \textit{statistical significance} of the returned patterns. The multiplicative approach shows more moderate improvements, under the MST or MLS pattern quality functions. It nevertheless yields high gains in variable correlation. Notably, the modified objective functions did not strongly alter pattern quality. Interestingly, in some instances, the MOF contributed to an increase in the average pattern quality, as seen in the case of the $\delta$-Trimax algorithm.


Considering the \textit{basketball} data, the results also show an overall improvement in \textit{discriminative power} criteria for Trigen. The multiplicative MOF with TriGen yields consistent gains of \textit{discriminative power} under both MSR or LSL pattern quality measurements, and, TriGen using an additive approach under the MLS measurement. As with the activity recognition data, gains of variable correlation were also observed when TriGen used a multiplicative approach with MSR as a pattern quality measure. Additionally, in both MOF versions of the $\delta$-Trimax algorithm, patterns achieved their maximum standard lift value but a lower lift value than the original approach. Both MOF versions of TriGen, under the LSL pattern quality measurement, and $\delta$-Trimax had significant gains in \textit{statistical significance}.



Finally, consider the results produced using the \textit{sports} data. We observe an increase of \textit{discriminative power} in all MOF versions of the TriGen algorithm versus its' original approach. Similarly to both activity recognition and basketball data results, results produced by the $\delta$-Trimax algorithm produced patterns with higher completeness of \textit{discriminative power}. The MOF multiplicative approach of TriGen, under the MLS and LSL pattern quality measurement, produced significant improvements in pattern \textit{statistical significance}. Similarly, both MOF versions of $\delta$-Trimax also produced patterns with higher \textit{statistical significance}.

\begin{figure}[!t]
\begin{adjustwidth}{-2.7cm}{-1.8cm}
\centering
\scriptsize
\includegraphics[height=3in]{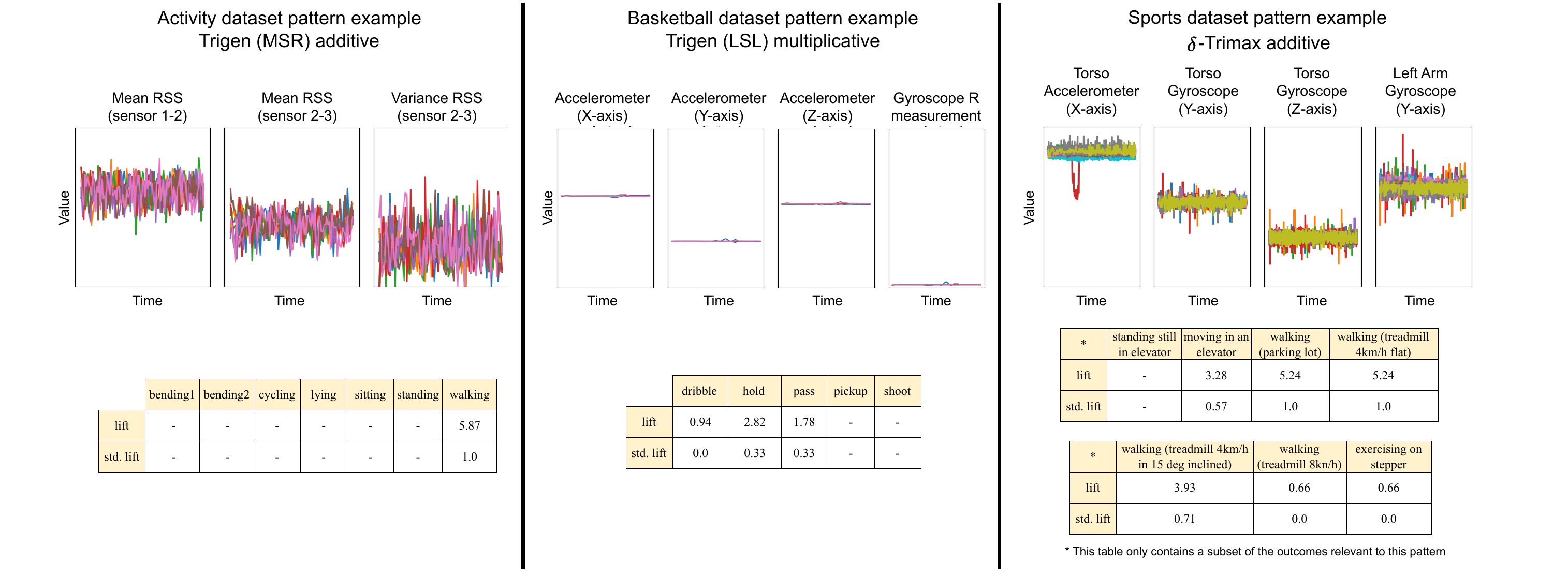}\\
\end{adjustwidth}

\caption{\small Patterns and their corresponding discriminative power on outcomes of interest. For each pattern, each line represents how one of the variables in the pattern varies across time, with the x-axis representing the passage of time and the y-axis the measurement. The pattern example of the activity dataset was extracted by TriGen (MSR) with the additive approach. The pattern example of the Basketball dataset was extracted by TriGen (LSL) using the multiplicative approach. Finally, the pattern example for the sports dataset was extracted by $\delta$-Trimax using the additive approach. Variable description: \\\footnotesize
\noindent \textbf{Activity dataset} \cite{misc_activity_recognition_system_based_on_multisensor_data_fusion}:  \\
-- Mean RSS (sensor 1-2) - mean received signal strength (RSS) between sensors 1 and 2; \\
-- Mean RSS (sensor 2-3) - mean received signal strength (RSS) between sensors 2 and 3; \\
-- Variance RSS (sensor 2-3) - variance received signal strength (RSS) between sensors 2 and 3. \\
-- R(m/$s^2$) - gyroscope measurement. \\
\noindent \textbf{Basketball dataset} \cite{misc_basketball_dataset_587}:  \\
-- Accelerometer (X-axis/Y-axis/Z-axis) - accelerometer measurement for specified axis; \\
-- Gyroscope R measurement - gyroscope measurement. \\
\noindent \textbf{Sports dataset} \cite{misc_daily_and_sports_activities_256}:  \\
-- Torso Accelerometer (X-axis) - torso accelerometer measurement for the x-axis; \\
-- Torso Gyroscope (Y-axis), Torso Gyroscope (Z-axis) - torso gyroscopes measurements for the y and x-axis; \\
-- Left Arm Gyroscope (Y-axis) - left arm gyroscope measurements for the y-axis. \\
}
\label{specific_pattern_example}
\end{figure}

In addition to the previous analysis, we provide an in-depth view of the discriminative power of patterns for each dataset. Figure \ref{specific_pattern_example} displays three distinct patterns. The first pattern is extracted from the \textit{activity} dataset using the MOF additive version of TriGen, assessed under the MSR pattern quality measurement. The second pattern originates from the \textit{basketball} dataset, extracted by the MOF multiplicative version of TriGen, and evaluated under the LSL pattern quality measurement. Lastly, the third pattern is obtained from the \textit{sports} dataset using the MOF additive version of $\delta$-Trimax. All of the patterns presented are from pattern solutions whose pattern exhibited significant improvements in \textit{discriminative power}. We note that the \textit{activity} dataset pattern only discriminates 'walking.' In the \textit{basketball} dataset, the pattern strongly discriminates 'hold' but also 'pass.' The selected pattern in sports data exhibits high discriminative power across walking-related tasks, most likely due to the similarity between tasks.

\newgeometry{top=2cm, bottom=2cm}

\begin{sidewaystable}

\caption{\small Statistics of the pattern solutions extracted by each triclustering algorithm version (original, additive-based, multiplicative-based), with different pattern quality metrics, over the three case study datasets. Column ''$|\overline{I}| \times |\overline{J}| \times |\overline{K}|$" presents details pertaining to the average number of observations ($|\overline{I}|$), variables ($|\overline{J}|$), and contexts ($|\overline{K}|$). The remaining columns correspond to the average$\hspace{0.1cm}\pm\hspace{0.1cm}$standard deviation of the measurement corresponding to the column name.}
\tiny
\begin{tabular}{p{0.8cm}p{2.2cm}|p{1.35cm}|p{1.6cm}p{0.9cm}p{1.2cm}|p{1.2cm}p{1.2cm}p{1.2cm}p{1.7cm}p{1.2cm}p{1.4cm}|p{1.2cm}p{1.2cm}}
& \multicolumn{13}{c}{\textbf{a) Activity dataset}} \\ \cmidrule{3-14} 
 & & $\overline{|I|}$$\times$$\overline{|J|}$$\times$$\overline{|K|}$ & MSR & MLS  & LSL  & \textit{lift} & \mbox{\textit{standard lift}} & DPC & SSC  & Pearson & Spearman & $MOF_{add.}$ & $MOF_{mul.}$ \\ \midrule
\multirow{9}{*}{Trigen} & Original (MSR) & 18 x 5 x 90  & 0.01$\hspace{0.1cm}\pm\hspace{0.1cm}$0.003   & -  & -  & 2.48$\hspace{0.1cm}\pm\hspace{0.1cm}$0.9   & 0.45$\hspace{0.1cm}\pm\hspace{0.1cm}$0.23 & 0.55$\hspace{0.1cm}\pm\hspace{0.1cm}$0.20 &  0.004$\hspace{0.1cm}\pm\hspace{0.1cm}$0.01 & 0.25$\hspace{0.1cm}\pm\hspace{0.1cm}$0.07  & 0.25$\hspace{0.1cm}\pm\hspace{0.1cm}$0.06  & -  & - \\ 
   & Additive (MSR) & 16 x 5 x 97  & 0.01$\hspace{0.1cm}\pm\hspace{0.1cm}$0.004   & -  & -  & \textbf{4.4$\hspace{0.1cm}\pm\hspace{0.1cm}$2.04}   & \textbf{0.75$\hspace{0.1cm}\pm\hspace{0.1cm}$0.23} & \textbf{0.29$\hspace{0.1cm}\pm\hspace{0.1cm}$0.18}   &  \textbf{0.003$\hspace{0.1cm}\pm\hspace{0.1cm}$0.0005}   & 0.24$\hspace{0.1cm}\pm\hspace{0.1cm}$0.11  & 0.24$\hspace{0.1cm}\pm\hspace{0.1cm}$0.12  & \textbf{0.40$\hspace{0.1cm}\pm\hspace{0.1cm}$0.06}  & -   \\ 
   & \mbox{Multiplicative (MSR)} & 14 x 4 x 86  & \textbf{0.01$\hspace{0.1cm}\pm\hspace{0.1cm}$0.003}   & -  & -  & \textbf{3.52$\hspace{0.1cm}\pm\hspace{0.1cm}$1.60}  & 0.54$\hspace{0.1cm}\pm\hspace{0.1cm}$0.23 & \textbf{0.44$\hspace{0.1cm}\pm\hspace{0.1cm}$0.20}   &  0.004$\hspace{0.1cm}\pm\hspace{0.1cm}$0.0009   & \textbf{0.45$\hspace{0.1cm}\pm\hspace{0.1cm}$0.19}  & \textbf{0.42$\hspace{0.1cm}\pm\hspace{0.1cm}$0.19}  & - & \textbf{0.02$\hspace{0.1cm}\pm\hspace{0.1cm}$0.06} \\ 
   & Original (MLS) & 13 x 5 x 92  & \multicolumn{1}{c}{-} & 0.48$\hspace{0.1cm}\pm\hspace{0.1cm}$0.07 & -  & 2.85$\hspace{0.1cm}\pm\hspace{0.1cm}$1.43  & 0.35$\hspace{0.1cm}\pm\hspace{0.1cm}$0.18 & 0.58$\hspace{0.1cm}\pm\hspace{0.1cm}$0.17  &  0.005$\hspace{0.1cm}\pm\hspace{0.1cm}$0.001   & 0.27$\hspace{0.1cm}\pm\hspace{0.1cm}$0.06  & 0.25$\hspace{0.1cm}\pm\hspace{0.1cm}$0.06 & - & - \\ 
   & Additive (MLS) & 14 x 4 x 89  & \multicolumn{1}{c}{-} & 0.49$\hspace{0.1cm}\pm\hspace{0.1cm}$0.04 & -  & \textbf{4.49$\hspace{0.1cm}\pm\hspace{0.1cm}$1.24}  & \textbf{0.76$\hspace{0.1cm}\pm\hspace{0.1cm}$0.19} & \textbf{0.26$\hspace{0.1cm}\pm\hspace{0.1cm}$0.12} &  \textbf{0.004$\hspace{0.1cm}\pm\hspace{0.1cm}$0.0009}  & 0.26$\hspace{0.1cm}\pm\hspace{0.1cm}$0.10  & 0.26$\hspace{0.1cm}\pm\hspace{0.1cm}$0.11  & \textbf{1.4$\hspace{0.1cm}\pm\hspace{0.1cm}$0.09} & - \\ 
   & \mbox{Multiplicative (MLS)} & 16 x 3 x 83  & \multicolumn{1}{c}{-} & 0.51$\hspace{0.1cm}\pm\hspace{0.1cm}$0.07 & -  & 3.04$\hspace{0.1cm}\pm\hspace{0.1cm}$1.00  & \textbf{0.51$\hspace{0.1cm}\pm\hspace{0.1cm}$0.21} & \textbf{0.47$\hspace{0.1cm}\pm\hspace{0.1cm}$0.18} &  \textbf{0.004$\hspace{0.1cm}\pm\hspace{0.1cm}$0.0008}  & \textbf{0.36$\hspace{0.1cm}\pm\hspace{0.1cm}$0.15}  & \textbf{0.36$\hspace{0.1cm}\pm\hspace{0.1cm}$0.16}  & - & \textbf{0.09$\hspace{0.1cm}\pm\hspace{0.1cm}$0.01}   \\ 
   & Original (LSL) & 16 x 5 x 86  & \multicolumn{1}{c}{-} & -  & 0.67$\hspace{0.1cm}\pm\hspace{0.1cm}$0.22 & 2.98$\hspace{0.1cm}\pm\hspace{0.1cm}$1.24  & 0.48$\hspace{0.1cm}\pm\hspace{0.1cm}$0.18 & 0.49$\hspace{0.1cm}\pm\hspace{0.1cm}$0.16 &  0.005$\hspace{0.1cm}\pm\hspace{0.1cm}$0.001  & 0.31$\hspace{0.1cm}\pm\hspace{0.1cm}$0.13  & 0.31$\hspace{0.1cm}\pm\hspace{0.1cm}$0.14  & - & -  \\ 
   & Additive (LSL) & 16 x 4 x 86  & \multicolumn{1}{c}{-} & -  & 0.57$\hspace{0.1cm}\pm\hspace{0.1cm}$0.21 & \textbf{4.18$\hspace{0.1cm}\pm\hspace{0.1cm}$1.20}  & \textbf{0.78$\hspace{0.1cm}\pm\hspace{0.1cm}$0.22} & \textbf{0.27$\hspace{0.1cm}\pm\hspace{0.1cm}$0.15} &  \textbf{0.004$\hspace{0.1cm}\pm\hspace{0.1cm}$0.0009}  & 0.34$\hspace{0.1cm}\pm\hspace{0.1cm}$0.16  & 0.33$\hspace{0.1cm}\pm\hspace{0.1cm}$0.16   & \textbf{1.42$\hspace{0.1cm}\pm\hspace{0.1cm}$0.27} & -  \\ 
   & Multivative (LSL) & 15 x 4 x 83  & \multicolumn{1}{c}{-} & -  & 0.60$\hspace{0.1cm}\pm\hspace{0.1cm}$0.13 & 3.33$\hspace{0.1cm}\pm\hspace{0.1cm}$1.33  & 0.55$\hspace{0.1cm}\pm\hspace{0.1cm}$0.24 & 0.44$\hspace{0.1cm}\pm\hspace{0.1cm}$0.20  &   0.004$\hspace{0.1cm}\pm\hspace{0.1cm}$0.001 & 0.33$\hspace{0.1cm}\pm\hspace{0.1cm}$0.13  & 0.33$\hspace{0.1cm}\pm\hspace{0.1cm}$0.13 & - & \textbf{0.09$\hspace{0.1cm}\pm\hspace{0.1cm}$0.02}   \\ \midrule
\multirow{3}{*}{$\delta$-Trimax} & Original & 32 x 5 x 50  & 0.009$\hspace{0.1cm}\pm\hspace{0.1cm}$0.0008 & -  & -  & 2.18$\hspace{0.1cm}\pm\hspace{0.1cm}$0.35  & 0.75$\hspace{0.1cm}\pm\hspace{0.1cm}$0.19 & 0.41$\hspace{0.1cm}\pm\hspace{0.1cm}$0.1 &  0.11$\hspace{0.1cm}\pm\hspace{0.1cm}$0.30  & 0.40$\hspace{0.1cm}\pm\hspace{0.1cm}$0.12  & 0.39$\hspace{0.1cm}\pm\hspace{0.1cm}$0.14 & - & -  \\ 
   & Additive & 27 x 4 x 37  & \textbf{0.004$\hspace{0.1cm}\pm\hspace{0.1cm}$0.001}  & -  & -  & 2.42$\hspace{0.1cm}\pm\hspace{0.1cm}$0.67  & 0.70$\hspace{0.1cm}\pm\hspace{0.1cm}$0.24 & 0.41$\hspace{0.1cm}\pm\hspace{0.1cm}$0.13  &  0.01$\hspace{0.1cm}\pm\hspace{0.1cm}$0.03 & 0.44$\hspace{0.1cm}\pm\hspace{0.1cm}$0.23  & 0.40$\hspace{0.1cm}\pm\hspace{0.1cm}$0.21 & \textbf{0.30$\hspace{0.1cm}\pm\hspace{0.1cm}$0.01} & - \\ 
   & Multiplicative & 28 x 4 x 71  & \textbf{0.008$\hspace{0.1cm}\pm\hspace{0.1cm}$0.001}  & -  & -  & 2.45$\hspace{0.1cm}\pm\hspace{0.1cm}$0.42  & 0.74$\hspace{0.1cm}\pm\hspace{0.1cm}$0.12 & 0.38$\hspace{0.1cm}\pm\hspace{0.1cm}$0.06   &  0.01$\hspace{0.1cm}\pm\hspace{0.1cm}$0.01   & 0.35$\hspace{0.1cm}\pm\hspace{0.1cm}$0.14  & 0.35$\hspace{0.1cm}\pm\hspace{0.1cm}$0.15 & - & \textbf{0.02$\hspace{0.1cm}\pm\hspace{0.1cm}$0.06}  \\ \midrule \\ 
 & \multicolumn{13}{c}{\textbf{b) Basketball dataset}}  \\ \cmidrule{3-14} 
 & & $\overline{|I|}$$\times$$\overline{|J|}$$\times$$\overline{|K|}$ & MSR & MLS   & LSL   & Lift & Std. Lift & DPC & SSC & Pearson & Spearman & $MOF_{add.}$ & $MOF_{mul.}$ \\ \midrule
\multirow{9}{*}{Trigen} & Original (MSR) & 17 x 6 x 22  & 0.009$\hspace{0.1cm}\pm\hspace{0.1cm}$0.003  & -  & -  & 1.67$\hspace{0.1cm}\pm\hspace{0.1cm}$0.35  & 0.31$\hspace{0.1cm}\pm\hspace{0.1cm}$0.06 & 0.72$\hspace{0.1cm}\pm\hspace{0.1cm}$0.09   &  0.08$\hspace{0.1cm}\pm\hspace{0.1cm}$0.21   & 0.19$\hspace{0.1cm}\pm\hspace{0.1cm}$0.06  & 0.20$\hspace{0.1cm}\pm\hspace{0.1cm}$0.06 &   - & -  \\ 
   & Additive (MSR) & 17 x 6 x 22  & 0.008$\hspace{0.1cm}\pm\hspace{0.1cm}$0.003  & -  & -  & 1.79$\hspace{0.1cm}\pm\hspace{0.1cm}$0.35  & \textbf{0.35$\hspace{0.1cm}\pm\hspace{0.1cm}$0.09} & \textbf{0.67$\hspace{0.1cm}\pm\hspace{0.1cm}$0.1} &  0.02$\hspace{0.1cm}\pm\hspace{0.1cm}$0.008   & 0.19$\hspace{0.1cm}\pm\hspace{0.1cm}$0.08  & 0.19$\hspace{0.1cm}\pm\hspace{0.1cm}$0.08  &  0.42$\hspace{0.1cm}\pm\hspace{0.1cm}$0.06 & -  \\ 
   & \mbox{Multiplicative (MSR)} & 13 x 4 x 22  & 0.006$\hspace{0.1cm}\pm\hspace{0.1cm}$0.005  & -  & -  & \textbf{2.36$\hspace{0.1cm}\pm\hspace{0.1cm}$0.87}  & \textbf{0.47$\hspace{0.1cm}\pm\hspace{0.1cm}$0.18} & \textbf{0.56$\hspace{0.1cm}\pm\hspace{0.1cm}$0.19} &  0.013$\hspace{0.1cm}\pm\hspace{0.1cm}$0.006  & \textbf{0.27$\hspace{0.1cm}\pm\hspace{0.1cm}$0.17}  & \textbf{0.26$\hspace{0.1cm}\pm\hspace{0.1cm}$0.14}   & - & \textbf{0.03$\hspace{0.1cm}\pm\hspace{0.1cm}$0.12}   \\ 
   & Original (MLS) & 13 x 5 x 20  & \multicolumn{1}{c}{-} & 0.53$\hspace{0.1cm}\pm\hspace{0.1cm}$0.04 & -  & 1.79$\hspace{0.1cm}\pm\hspace{0.1cm}$0.40  & 0.30$\hspace{0.1cm}\pm\hspace{0.1cm}$0.11 & 0.70$\hspace{0.1cm}\pm\hspace{0.1cm}$0.11 &  0.05$\hspace{0.1cm}\pm\hspace{0.1cm}$0.05  & 0.22$\hspace{0.1cm}\pm\hspace{0.1cm}$0.15  & 0.23$\hspace{0.1cm}\pm\hspace{0.1cm}$0.10 & - & -  \\ 
   & Additive (MLS) & 10 x 5 x 18  & \multicolumn{1}{c}{-} & 0.54$\hspace{0.1cm}\pm\hspace{0.1cm}$0.04 & -  & \textbf{3.17$\hspace{0.1cm}\pm\hspace{0.1cm}$0.83}  & \textbf{0.62$\hspace{0.1cm}\pm\hspace{0.1cm}$0.22} & \textbf{0.39$\hspace{0.1cm}\pm\hspace{0.1cm}$0.16}   &  0.03$\hspace{0.1cm}\pm\hspace{0.1cm}$0.02   & 0.24$\hspace{0.1cm}\pm\hspace{0.1cm}$0.13  & 0.26$\hspace{0.1cm}\pm\hspace{0.1cm}$0.16  & \textbf{1.65$\hspace{0.1cm}\pm\hspace{0.1cm}$0.10} & - \\ 
   & \mbox{Multiplicative (MLS)} & 11 x 3 x 20  & \multicolumn{1}{c}{-} & 0.58$\hspace{0.1cm}\pm\hspace{0.1cm}$0.05 & -  & \textbf{2.19$\hspace{0.1cm}\pm\hspace{0.1cm}$0.63}  & 0.34$\hspace{0.1cm}\pm\hspace{0.1cm}$0.14 & 0.62$\hspace{0.1cm}\pm\hspace{0.1cm}$0.13   &  \textbf{0.13$\hspace{0.1cm}\pm\hspace{0.1cm}$0.006}   & 0.30$\hspace{0.1cm}\pm\hspace{0.1cm}$0.18  & 0.30$\hspace{0.1cm}\pm\hspace{0.1cm}$0.16  & - & \textbf{0.16$\hspace{0.1cm}\pm\hspace{0.1cm}$0.02} \\ 
   & Original (LSL) & 14 x 4 x 17  & \multicolumn{1}{c}{-} & -  & 0.45$\hspace{0.1cm}\pm\hspace{0.1cm}$0.13 & 1.80$\hspace{0.1cm}\pm\hspace{0.1cm}$0.38  & 0.32$\hspace{0.1cm}\pm\hspace{0.1cm}$0.11 & 0.69$\hspace{0.1cm}\pm\hspace{0.1cm}$0.13 &  0.22$\hspace{0.1cm}\pm\hspace{0.1cm}$0.33  & 0.33$\hspace{0.1cm}\pm\hspace{0.1cm}$0.16  & 0.31$\hspace{0.1cm}\pm\hspace{0.1cm}$0.16 & - & -  \\ 
   & Additive (LSL) & 14 x 4 x 20  & \multicolumn{1}{c}{-} & -  & 0.45$\hspace{0.1cm}\pm\hspace{0.1cm}$0.13 & 1.96$\hspace{0.1cm}\pm\hspace{0.1cm}$0.46  & 0.36$\hspace{0.1cm}\pm\hspace{0.1cm}$0.10 & 0.64$\hspace{0.1cm}\pm\hspace{0.1cm}$0.11 &  \textbf{0.02$\hspace{0.1cm}\pm\hspace{0.1cm}$0.01}  & 0.37$\hspace{0.1cm}\pm\hspace{0.1cm}$0.16  & 0.34$\hspace{0.1cm}\pm\hspace{0.1cm}$0.17 & \textbf{1.64$\hspace{0.1cm}\pm\hspace{0.1cm}$0.13} &  -   \\ 
   & Multivative (LSL) & 11 x 3 x 20  & \multicolumn{1}{c}{-} & -  & 0.61$\hspace{0.1cm}\pm\hspace{0.1cm}$0.16 & \textbf{2.62$\hspace{0.1cm}\pm\hspace{0.1cm}$0.84}  & \textbf{0.48$\hspace{0.1cm}\pm\hspace{0.1cm}$0.23} & \textbf{0.52$\hspace{0.1cm}\pm\hspace{0.1cm}$0.19}  &  \textbf{0.02$\hspace{0.1cm}\pm\hspace{0.1cm}$0.007} & 0.33$\hspace{0.1cm}\pm\hspace{0.1cm}$0.22  & 0.33$\hspace{0.1cm}\pm\hspace{0.1cm}$0.23  &  - & \textbf{0.15$\hspace{0.1cm}\pm\hspace{0.1cm}$0.03}  \\ \midrule
\multirow{3}{*}{$\delta$-Trimax} & Original & 15 x 3 x 15  & 0.0009$\hspace{0.1cm}\pm\hspace{0.1cm}$0.0004   & -  & -  & 2.77$\hspace{0.1cm}\pm\hspace{0.1cm}$0.43  & 0.57$\hspace{0.1cm}\pm\hspace{0.1cm}$0.21 & 0.44$\hspace{0.1cm}\pm\hspace{0.1cm}$0.13 &  0.36$\hspace{0.1cm}\pm\hspace{0.1cm}$0.46  & 0.52$\hspace{0.1cm}\pm\hspace{0.1cm}$0.30  & 0.52$\hspace{0.1cm}\pm\hspace{0.1cm}$0.28 & - & - \\ 
   & Additive & 38 x 5 x 20  & 0.003$\hspace{0.1cm}\pm\hspace{0.1cm}$0.0006 & -  & -  & 2.11$\hspace{0.1cm}\pm\hspace{0.1cm}$0.30  & \textbf{1.0$\hspace{0.1cm}\pm\hspace{0.1cm}$0.0}   & \textbf{0.29$\hspace{0.1cm}\pm\hspace{0.1cm}$0.04}  &  \textbf{0.02$\hspace{0.1cm}\pm\hspace{0.1cm}$0.02} & 0.34$\hspace{0.1cm}\pm\hspace{0.1cm}$0.12  & 0.34$\hspace{0.1cm}\pm\hspace{0.1cm}$0.11 & 1.52$\hspace{0.1cm}\pm\hspace{0.1cm}$0.07 & -  \\ 
   & Multiplicative & 53 x 6 x 43  & 0.004$\hspace{0.1cm}\pm\hspace{0.1cm}$0.0006 & -  & -  & 1.50$\hspace{0.1cm}\pm\hspace{0.1cm}$0.19  & \textbf{1.0$\hspace{0.1cm}\pm\hspace{0.1cm}$0.0}   & 0.41$\hspace{0.1cm}\pm\hspace{0.1cm}$0.04 &  \textbf{0.003$\hspace{0.1cm}\pm\hspace{0.1cm}$0.0003}  & 0.22$\hspace{0.1cm}\pm\hspace{0.1cm}$0.02  & 0.22$\hspace{0.1cm}\pm\hspace{0.1cm}$0.02 & - & \textbf{0.01$\hspace{0.1cm}\pm\hspace{0.1cm}$0.02}   \\ \midrule \\ 
 & \multicolumn{13}{c}{\textbf{c) Sports dataset}}   \\ \cmidrule{3-14} 
 & & $\overline{|I|}$$\times$$\overline{|J|}$$\times$$\overline{|K|}$ & MSR & MLS   & LSL   & Lift & Std. Lift & DPC & SSC & Pearson & Spearman & $MOF_{add.}$ & $MOF_{mul.}$ \\ \midrule
\multirow{9}{*}{Trigen} & Original (MSR) & 19 x 25 x 95 & 33.58$\hspace{0.1cm}\pm\hspace{0.1cm}$34.89  & -  & -  & 3.72$\hspace{0.1cm}\pm\hspace{0.1cm}$1.80  & 0.39$\hspace{0.1cm}\pm\hspace{0.1cm}$0.22 & 0.49$\hspace{0.1cm}\pm\hspace{0.1cm}$0.16   &  0.004$\hspace{0.1cm}\pm\hspace{0.1cm}$0.0006   & 0.03$\hspace{0.1cm}\pm\hspace{0.1cm}$0.01  & 0.03$\hspace{0.1cm}\pm\hspace{0.1cm}$0.01   & - & -   \\ 
   & Additive (MSR) & 16 x 21 x 90 & 37.81$\hspace{0.1cm}\pm\hspace{0.1cm}$53.94  & -  & -  & \textbf{7.82$\hspace{0.1cm}\pm\hspace{0.1cm}$4.88}  & \textbf{0.69$\hspace{0.1cm}\pm\hspace{0.1cm}$0.31} & \textbf{0.26$\hspace{0.1cm}\pm\hspace{0.1cm}$0.21}   &  0.004$\hspace{0.1cm}\pm\hspace{0.1cm}$0.0009   & \textbf{0.04$\hspace{0.1cm}\pm\hspace{0.1cm}$0.03}  & \textbf{0.04$\hspace{0.1cm}\pm\hspace{0.1cm}$0.03}  & 3.62$\hspace{0.1cm}\pm\hspace{0.1cm}$3.97 & -  \\ 
   & \mbox{Multiplicative (MSR)} & 15 x 18 x 89 & 35.92$\hspace{0.1cm}\pm\hspace{0.1cm}$60.10  & -  & -  & \textbf{7.92$\hspace{0.1cm}\pm\hspace{0.1cm}$4.82}  & \textbf{0.60$\hspace{0.1cm}\pm\hspace{0.1cm}$0.31} & \textbf{0.31$\hspace{0.1cm}\pm\hspace{0.1cm}$0.22} &  0.004$\hspace{0.1cm}\pm\hspace{0.1cm}$0.0007  & \textbf{0.07$\hspace{0.1cm}\pm\hspace{0.1cm}$0.11}  & \textbf{0.07$\hspace{0.1cm}\pm\hspace{0.1cm}$0.09} & - & 0.18$\hspace{0.1cm}\pm\hspace{0.1cm}$3.92  \\ 
   & Original (MLS) & 17 x 19 x 87 & \multicolumn{1}{c}{-} & 0.50$\hspace{0.1cm}\pm\hspace{0.1cm}$0.01 & -  & 4.72$\hspace{0.1cm}\pm\hspace{0.1cm}$2.37  & 0.47$\hspace{0.1cm}\pm\hspace{0.1cm}$0.31 & 0.43$\hspace{0.1cm}\pm\hspace{0.1cm}$0.23  &  0.004$\hspace{0.1cm}\pm\hspace{0.1cm}$0.001 & 0.05$\hspace{0.1cm}\pm\hspace{0.1cm}$0.03  & 0.05$\hspace{0.1cm}\pm\hspace{0.1cm}$0.02  & - &  - \\ 
   & Additive (MLS) & 13 x 19 x 85 & \multicolumn{1}{c}{-} & 0.50$\hspace{0.1cm}\pm\hspace{0.1cm}$0.01 & -  & \textbf{11.20$\hspace{0.1cm}\pm\hspace{0.1cm}$4.72} & \textbf{0.86$\hspace{0.1cm}\pm\hspace{0.1cm}$0.22} & \textbf{0.14$\hspace{0.1cm}\pm\hspace{0.1cm}$0.14}   &  0.004$\hspace{0.1cm}\pm\hspace{0.1cm}$0.0008   & 0.05$\hspace{0.1cm}\pm\hspace{0.1cm}$0.03  & 0.05$\hspace{0.1cm}\pm\hspace{0.1cm}$0.03  & \textbf{1.28$\hspace{0.1cm}\pm\hspace{0.1cm}$0.12} & - \\ 
   & \mbox{Multiplicative (MLS)} & 15 x 14 x 91 & \multicolumn{1}{c}{-} & 0.49$\hspace{0.1cm}\pm\hspace{0.1cm}$0.02 & -  & \textbf{7.59$\hspace{0.1cm}\pm\hspace{0.1cm}$3.96}  & 0.63$\hspace{0.1cm}\pm\hspace{0.1cm}$0.31 & \textbf{0.29$\hspace{0.1cm}\pm\hspace{0.1cm}$0.21} &  \textbf{0.003$\hspace{0.1cm}\pm\hspace{0.1cm}$0.0003}  & \textbf{0.09$\hspace{0.1cm}\pm\hspace{0.1cm}$0.07}  & \textbf{0.10$\hspace{0.1cm}\pm\hspace{0.1cm}$0.07}  & - & \textbf{0.07$\hspace{0.1cm}\pm\hspace{0.1cm}$0.02} \\ 
   & Original (LSL) & 18 x 20 x 88 & \multicolumn{1}{c}{-} & -  & 0.57$\hspace{0.1cm}\pm\hspace{0.1cm}$0.10 & 3.93$\hspace{0.1cm}\pm\hspace{0.1cm}$1.33  & 0.39$\hspace{0.1cm}\pm\hspace{0.1cm}$0.22 & 0.47$\hspace{0.1cm}\pm\hspace{0.1cm}$0.14 &  0.004$\hspace{0.1cm}\pm\hspace{0.1cm}$0.001  & 0.04$\hspace{0.1cm}\pm\hspace{0.1cm}$0.03  & 0.05$\hspace{0.1cm}\pm\hspace{0.1cm}$0.03 & - & -  \\ 
   & Additive (LSL) & 14 x 16 x 77 & \multicolumn{1}{c}{-} & -  & 0.52$\hspace{0.1cm}\pm\hspace{0.1cm}$0.14 & \textbf{10.34$\hspace{0.1cm}\pm\hspace{0.1cm}$4.28} & \textbf{0.84$\hspace{0.1cm}\pm\hspace{0.1cm}$0.21} & \textbf{0.15$\hspace{0.1cm}\pm\hspace{0.1cm}$0.13}   &  0.004$\hspace{0.1cm}\pm\hspace{0.1cm}$0.002   & 0.05$\hspace{0.1cm}\pm\hspace{0.1cm}$0.03  & 0.06$\hspace{0.1cm}\pm\hspace{0.1cm}$0.04   & \textbf{1.31$\hspace{0.1cm}\pm\hspace{0.1cm}$0.14} & -   \\ 
   & Multivative (LSL) & 15 x 16 x 94 & \multicolumn{1}{c}{-} & -  & 0.56$\hspace{0.1cm}\pm\hspace{0.1cm}$0.12 & \textbf{7.99$\hspace{0.1cm}\pm\hspace{0.1cm}$5.33}  & \textbf{0.62$\hspace{0.1cm}\pm\hspace{0.1cm}$0.29} & \textbf{0.30$\hspace{0.1cm}\pm\hspace{0.1cm}$0.20} &  \textbf{0.003$\hspace{0.1cm}\pm\hspace{0.1cm}$0.0003}  & 0.06$\hspace{0.1cm}\pm\hspace{0.1cm}$0.03  & 0.06$\hspace{0.1cm}\pm\hspace{0.1cm}$0.03  & - & \textbf{0.07$\hspace{0.1cm}\pm\hspace{0.1cm}$0.02} \\ \midrule
\multirow{3}{*}{$\delta$-Trimax} & Original & 9 x 6 x 63 & $9$$\times$$10^{-5}$$\pm$$7$$\times$$10^{-6}$   & -  & -  & 7.45$\hspace{0.1cm}\pm\hspace{0.1cm}$2.06  & 0.36$\hspace{0.1cm}\pm\hspace{0.1cm}$0.14   & 0.40$\hspace{0.1cm}\pm\hspace{0.1cm}$0.09   &  0.50$\hspace{0.1cm}\pm\hspace{0.1cm}$0.49   & 0.27$\hspace{0.1cm}\pm\hspace{0.1cm}$0.15  & 0.27$\hspace{0.1cm}\pm\hspace{0.1cm}$0.17 & - &  - \\ 
   & Additive & 30 x 7 x 326   & $6$$\times$$10^{-4}$$\pm$$2$$\times$$10^{-4}$ & -  & -  & 4.67$\hspace{0.1cm}\pm\hspace{0.1cm}$0.38   & \textbf{0.91$\hspace{0.1cm}\pm\hspace{0.1cm}$0.06}   & \textbf{0.17$\hspace{0.1cm}\pm\hspace{0.1cm}$0.04}   &  \textbf{0.003$\hspace{0.1cm}\pm\hspace{0.1cm}$9$\times$$10^{-7}$} & 0.19$\hspace{0.1cm}\pm\hspace{0.1cm}$0.05   & 0.20$\hspace{0.1cm}\pm\hspace{0.1cm}$0.05 & 0.14$\hspace{0.1cm}\pm\hspace{0.1cm}$0.01 & -   \\ 
   & Multiplicative & \mbox{41 x 45 x 372}   & 1.03$\hspace{0.1cm}\pm\hspace{0.1cm}$1.02  & -  & -  & 3.81$\hspace{0.1cm}\pm\hspace{0.1cm}$0.65  & 1.0$\hspace{0.1cm}\pm\hspace{0.1cm}$0.0   & \textbf{0.16$\hspace{0.1cm}\pm\hspace{0.1cm}$0.02}  &  \textbf{0.003$\hspace{0.1cm}\pm\hspace{0.1cm}$$ 9$$\times$$10^{-7}$} & \textbf{0.01}$\hspace{0.1cm}\pm\hspace{0.1cm}$6$\times$$10^{-4}$ & 0.01$\hspace{0.1cm}\pm\hspace{0.1cm}$0.0002 & - & 0.05$\hspace{0.1cm}\pm\hspace{0.1cm}$0.86 \\ \midrule
\end{tabular}
\label{tab_results}
\end{sidewaystable}
\restoregeometry

\section{Conclusion}

This work proposed a novel approach for extracting patterns from tensor data, ensuring guarantees of pattern quality, statistical significance, and discriminative power. Firstly, we provided an in-depth discussion on the challenges faced by different greedy-based approaches and how to tackle them. Secondly, we provided sound statistical reasoning on how to parameterize threshold-based greedy approaches to accommodate the proposed modified objective functions. Finally, we showed the effectiveness of our methodology by testing it in the context of three case studies. Pattern solutions extracted by the modified versions of the algorithms had significant improvements in pattern \textit{discriminative power}, pattern \textit{statistical significance}, and in some cases, higher pattern quality, against the classic quality-driven solutions produced by the original approaches.

While our methodology reveals notable quantitative and qualitative improvements against the original approaches, we acknowledge that the importance of each criterion may vary between domains. In this context, the hyperparametization of the integrative scoring function should be considered. 
Furthermore, the inclusion of domain relevance as a fourth criteria class can be pursued to enhance the actionability of patterns without forgoing the established improvements. Finally, as our methodology can be straightforwardly extended towards the discovery of patterns in other data structures, a broader assessment of its impact for knowledge acquisition from tabular and $N$-way tensor (e.g., $N>4$) data structures is highlighted for future work.

\subsection*{Funding}
\small This work was supported by Fundação para a Ciência e Tecnologia (FCT) under contract CEECIND/01399/ 2017/CP1462/CT0015 to RSC, FCT individual PhD grant to LA (2021.07759.BD), INESC-ID plurianual (UIDB/50021/2020), and further supported by the Associate Laboratory for Green Chemistry (LAQV), financed by national funds from FCT/MCTES (UIDB/50006/2020 and UIDP/50006/2020).

\footnotesize
\bibliographystyle{plain}
\bibliography{references}   

\begin{thebibliography}{10}

\bibitem{ahmed2014shifting}
Hasin~Afzal Ahmed, Priyakshi Mahanta, Dhruba~Kumar Bhattacharyya, and
  Jugal~Kumar Kalita.
\newblock Shifting-and-scaling correlation based biclustering algorithm.
\newblock {\em IEEE/ACM transactions on computational biology and
  bioinformatics}, 11(6):1239--1252, 2014.

\bibitem{alexandre2022disa}
Leonardo Alexandre, Rafael~S Costa, and Rui Henriques.
\newblock Disa tool: Discriminative and informative subspace assessment with
  categorical and numerical outcomes.
\newblock {\em Plos one}, 17(10):e0276253, 2022.

\bibitem{alexandre2023trisig}
Leonardo Alexandre, Rafael~S Costa, and Rui Henriques.
\newblock Trisig: Assessing the statistical significance of triclusters.
\newblock {\em arXiv preprint arXiv:2306.00643}, 2023.

\bibitem{alexandre2021mining}
Leonardo Alexandre, Rafael~S Costa, L{\'u}cio~Lara Santos, and Rui Henriques.
\newblock Mining pre-surgical patterns able to discriminate post-surgical
  outcomes in the oncological domain.
\newblock {\em IEEE Journal of Biomedical and Health Informatics},
  25(7):2421--2434, 2021.

\bibitem{amaro2021generating}
Jos{\'e}~L Amaro-Mellado, Laura Melgar-Garc{\'\i}a, Cristina Rubio-Escudero,
  and David Guti{\'e}rrez-Avil{\'e}s.
\newblock Generating a seismogenic source zone model for the pyrenees: A
  gis-assisted triclustering approach.
\newblock {\em Computers \& Geosciences}, 150:104736, 2021.

\bibitem{misc_basketball_dataset_587}
Unknown Author.
\newblock {Basketball dataset}.
\newblock UCI Machine Learning Repository, 2019.
\newblock {DOI}: https://doi.org/10.24432/C56G77.

\bibitem{misc_daily_and_sports_activities_256}
Billur Barshan and Kerem Altun.
\newblock {Daily and Sports Activities}.
\newblock UCI Machine Learning Repository, 2013.
\newblock {DOI}: https://doi.org/10.24432/C5C59F.

\bibitem{bhar2013coexpression}
Anirban Bhar, Martin Haubrock, Anirban Mukhopadhyay, Ujjwal Maulik, Sanghamitra
  Bandyopadhyay, and Edgar Wingender.
\newblock Coexpression and coregulation analysis of time-series gene expression
  data in estrogen-induced breast cancer cell.
\newblock {\em Algorithms for molecular biology}, 8:1--11, 2013.

\bibitem{bhar2015multiobjective}
Anirban Bhar, Martin Haubrock, Anirban Mukhopadhyay, and Edgar Wingender.
\newblock Multiobjective triclustering of time-series transcriptome data
  reveals key genes of biological processes.
\newblock {\em BMC bioinformatics}, 16(1):1--19, 2015.

\bibitem{chen2016computational}
Bernard Chen, Christopher Rhodes, Alexander Yu, and Valentin Velchev.
\newblock The computational wine wheel 2.0 and the trimax triclustering in
  wineinformatics.
\newblock In {\em Advances in Data Mining. Applications and Theoretical
  Aspects: 16th Industrial Conference, ICDM 2016, New York, NY, USA, July
  13-17, 2016. Proceedings 16}, pages 223--238. Springer, 2016.

\bibitem{cheng2000biclustering}
Yizong Cheng and George~M Church.
\newblock Biclustering of expression data.
\newblock In {\em Ismb}, volume~8, pages 93--103, 2000.

\bibitem{gene25ashburner}
Gene~Ontology Consortium et~al.
\newblock Ashburner m, ball ca, blake ja, botstein d, butler h, cherry jm,
  davis ap, dolinski k, dwight ss, et al. 2000.
\newblock {\em Gene Ontology: tool for the unification of biology. Nat Genet},
  25:25--29, 2000.

\bibitem{gutierrez2014lsl}
David Guti{\'e}rrez-Avil{\'e}s and Cristina Rubio-Escudero.
\newblock Lsl: A new measure to evaluate triclusters.
\newblock In {\em 2014 IEEE International Conference on Bioinformatics and
  Biomedicine (BIBM)}, pages 30--37. IEEE, 2014.

\bibitem{gutierrez2015msl}
David Guti{\'e}rrez-Avil{\'e}s and Cristina Rubio-Escudero.
\newblock Msl: a measure to evaluate three-dimensional patterns in gene
  expression data.
\newblock {\em Evolutionary Bioinformatics}, 11:EBO--S25822, 2015.

\bibitem{gutierrez2016triq}
David Guti{\'e}rrez-Avil{\'e}s and Cristina Rubio-Escudero.
\newblock Triq: a comprehensive evaluation measure for triclustering
  algorithms.
\newblock In {\em Hybrid Artificial Intelligent Systems: 11th International
  Conference, HAIS 2016, Seville, Spain, April 18-20, 2016, Proceedings 11},
  pages 673--684. Springer, 2016.

\bibitem{gutierrez2014mining}
David Guti{\'e}rrez-Avil{\'e}s, Cristina Rubio-Escudero, et~al.
\newblock Mining 3d patterns from gene expression temporal data: a new
  tricluster evaluation measure.
\newblock {\em The Scientific World Journal}, 2014, 2014.

\bibitem{gutierrez2014trigen}
David Guti{\'e}rrez-Avil{\'e}s, Cristina Rubio-Escudero, Francisco
  Mart{\'\i}nez-{\'A}lvarez, and Jos{\'e}~C Riquelme.
\newblock Trigen: A genetic algorithm to mine triclusters in temporal gene
  expression data.
\newblock {\em Neurocomputing}, 132:42--53, 2014.

\bibitem{henriques2017bicpams}
Rui Henriques, Francisco~L Ferreira, and Sara~C Madeira.
\newblock Bicpams: software for biological data analysis with pattern-based
  biclustering.
\newblock {\em BMC bioinformatics}, 18:1--16, 2017.

\bibitem{henriques2018bsig}
Rui Henriques and Sara~C Madeira.
\newblock Bsig: evaluating the statistical significance of biclustering
  solutions.
\newblock {\em Data Mining and Knowledge Discovery}, 32:124--161, 2018.

\bibitem{henriques2018triclustering}
Rui Henriques and Sara~C Madeira.
\newblock Triclustering algorithms for three-dimensional data analysis: a
  comprehensive survey.
\newblock {\em ACM Computing Surveys (CSUR)}, 51(5):1--43, 2018.

\bibitem{henriques2021flebic}
Rui Henriques and Sara~C Madeira.
\newblock Flebic: Learning classifiers from high-dimensional biomedical data
  using discriminative biclusters with non-constant patterns.
\newblock {\em Pattern Recognition}, 115:107900, 2021.

\bibitem{kakati2018thd}
Tulika Kakati, Hasin~A Ahmed, Dhruba~K Bhattacharyya, and Jugal~K Kalita.
\newblock Thd-tricluster: A robust triclustering technique and its application
  in condition specific change analysis in hiv-1 progression data.
\newblock {\em Computational biology and chemistry}, 75:154--167, 2018.

\bibitem{liu2008multi}
Junwan Liu, Zhoujun Li, Xiaohua Hu, and Yiming Chen.
\newblock Multi-objective evolutionary algorithm for mining 3d clusters in
  gene-sample-time microarray data.
\newblock In {\em 2008 IEEE International Conference on Granular Computing},
  pages 442--447. IEEE, 2008.

\bibitem{madeira2004biclustering}
Sara~C Madeira and Arlindo~L Oliveira.
\newblock Biclustering algorithms for biological data analysis: a survey.
\newblock {\em IEEE/ACM transactions on computational biology and
  bioinformatics}, 1(1):24--45, 2004.

\bibitem{mcnicholas2008standardising}
Paul~David McNicholas, Thomas~Brendan Murphy, and M~O’Regan.
\newblock Standardising the lift of an association rule.
\newblock {\em Computational Statistics \& Data Analysis}, 52(10):4712--4721,
  2008.

\bibitem{melgar2021discovering}
Laura Melgar-Garc{\'\i}a, David Guti{\'e}rrez-Avil{\'e}s, Cristina
  Rubio-Escudero, and Alicia Troncoso.
\newblock Discovering three-dimensional patterns in real-time from data
  streams: An online triclustering approach.
\newblock {\em Information Sciences}, 558:174--193, 2021.

\bibitem{mukhopadhyay2015survey}
Anirban Mukhopadhyay, Ujjwal Maulik, and Sanghamitra Bandyopadhyay.
\newblock A survey of multiobjective evolutionary clustering.
\newblock {\em ACM Computing Surveys (CSUR)}, 47(4):1--46, 2015.

\bibitem{misc_activity_recognition_system_based_on_multisensor_data_fusion}
{Palumbo, Filippo and Gallicchio, Claudio and Pucci, Rita, and Micheli,
  Alessio}.
\newblock {Activity Recognition system based on Multisensor data fusion
  (AReM)}.
\newblock UCI Machine Learning Repository, 2016.
\newblock {DOI}: https://doi.org/10.24432/C5SS33.

\bibitem{pellegrina2019spumante}
Leonardo Pellegrina, Matteo Riondato, and Fabio Vandin.
\newblock Spumante: Significant pattern mining with unconditional testing.
\newblock In {\em Proceedings of the 25th ACM SIGKDD International Conference
  on Knowledge Discovery \& Data Mining}, pages 1528--1538, 2019.

\bibitem{seyfi2023h}
Majid Seyfi and Yue Xu.
\newblock H-dac: discriminative associative classification in data streams.
\newblock {\em Soft Computing}, 27(2):953--971, 2023.

\bibitem{soares2022learning}
Diogo~F Soares, Rui Henriques, Marta Gromicho, Mamede de~Carvalho, and Sara~C
  Madeira.
\newblock Learning prognostic models using a mixture of biclustering and
  triclustering: Predicting the need for non-invasive ventilation in
  amyotrophic lateral sclerosis.
\newblock {\em Journal of Biomedical Informatics}, 134:104172, 2022.

\bibitem{tang2022mining}
Huijun Tang, Jiangbo Qian, Yangguang Liu, and Xiao-Zhi Gao.
\newblock Mining statistically significant patterns with high utility.
\newblock {\em International Journal of Computational Intelligence Systems},
  15(1):88, 2022.

\bibitem{xie2019time}
Juan Xie, Anjun Ma, Anne Fennell, Qin Ma, and Jing Zhao.
\newblock It is time to apply biclustering: a comprehensive review of
  biclustering applications in biological and biomedical data.
\newblock {\em Briefings in bioinformatics}, 20(4):1450--1465, 2019.

\end{thebibliography}

\end{document}